\documentclass[conference]{IEEEtran}
\IEEEoverridecommandlockouts

\usepackage{cite}
\usepackage{amsmath,amssymb,amsfonts}
\usepackage{algorithmic}
\usepackage{graphicx}
\usepackage{textcomp}
\usepackage{xcolor}
\usepackage{booktabs}
\usepackage{cleveref}
\def\BibTeX{{\rm B\kern-.05em{\sc i\kern-.025em b}\kern-.08em
    T\kern-.1667em\lower.7ex\hbox{E}\kern-.125emX}}
\ifCLASSOPTIONcompsoc
    \usepackage[caption=false, font=normalsize, labelfont=sf, textfont=sf]{subfig}
\else
\usepackage[caption=false, font=footnotesize]{subfig}
\fi

\begin{document}

\title{Semantic and episodic memories in a predictive coding model of the neocortex
}

\author{\IEEEauthorblockN{Lucie Fontaine \IEEEauthorrefmark{1}, Frédéric Alexandre \IEEEauthorrefmark{2}}
\IEEEauthorblockA{\textit{
Centre Inria de l’université de Bordeaux} \\
\textit{Laboratoire Bordelais de Recherche en Informatique (LaBRI)}\\
\textit{Institut des Maladies Neurodégénératives (IMN)}\\
Bordeaux, France \\
Email: \IEEEauthorrefmark{1} fontaluc@gmail.com, \IEEEauthorrefmark{2} frederic.alexandre@inria.fr}}

\maketitle

\begin{abstract}
Complementary Learning Systems theory holds that intelligent agents need two learning systems. Semantic memory is encoded in the neocortex with dense, overlapping representations and acquires structured knowledge. Episodic memory is encoded in the hippocampus with sparse, pattern-separated representations and quickly learns the specifics of individual experiences. Recently, this duality between semantic and episodic memories has been challenged by predictive coding, a biologically plausible neural network model of the neocortex which was shown to have hippocampus-like abilities on auto-associative memory tasks. These results raise the question of the episodic capabilities of the neocortex and their relation to semantic memory. In this paper, we present such a predictive coding model of the neocortex and explore its episodic capabilities. We show that this kind of model can indeed recall the specifics of individual examples but only if it is trained on a small number of examples. The model is overfitted to these exemples and does not generalize well, suggesting that episodic memory can arise from semantic learning. Indeed, a model trained with many more examples loses its recall capabilities. This work suggests that individual examples can be encoded gradually in the neocortex using dense, overlapping representations but only in a limited number, motivating the need for sparse, pattern-separated representations as found in the hippocampus. 
\end{abstract}


\section{Introduction}

Different models of neural networks have been shown to store different representations of information, that can be related in neuroscience to episodic memory and semantic memory \cite{spens2024generative, KáliSzabolcs2004Ormd}. The way these representations of information interact is much discussed both in Neuroscience and AI with tremendous implications in the nature of underlying cognitive functions \cite{kumaran2016learning}. This debate motivates the present experimental study. 

\subsection{Neuroscience}

\subsubsection{Complementary Learning Systems}

According to Complementary Learning Systems (CLS) theory, intelligent agents need episodic and semantic memories \cite{kumaran2016learning}. Episodic memory corresponds to an individual's sensory and emotional experience, whereas semantic memory is one's knowledge about the world. For example, the memory of the particular dog you saw yesterday is considered episodic, whereas the knowledge you have about how dogs in general look like would be semantic. In mammals, the hippocampus and neocortex are respectively thought to be mainly responsible for episodic and semantic memories \cite{kumaran2016learning}. In these brain regions, these two kinds of memory differ in their representations and storage. Whereas episodic memory involves sparse, non-overlapping (pattern-separated) representations which are acquired rapidly by the hippocampus, semantic memory involves dense, overlapping representations which are acquired gradually by the neocortex. Following up on our previous example, you can remember yesterday's dog after seeing it only once whereas knowledge about dogs is typically acquired after seeing multiple dogs. In the hippocampus, two dogs you have seen will be represented quite differently despite their similarity to avoid interference, whereas in the neocortex they will be represented similarly to allow generalization.

According to the CLS theory, episodic memories are transformed into semantic ones through hippocampal replay. This division into two learning sytems is assumed to be a solution to the problem of catastrophic forgetting or interference. When an agent learns a task and switches to a new task, replays of the experience of the first task are interleaved with examples of the second task to avoid forgetting the old task.

\subsubsection{Consolidation theory} 

In contrast, in the standard consolidation theory (SCT), episodic memories are simply transferred to the neocortex through consolidation \cite{SQUIRE1995169}. It builds upon the observation that patients with hippocampal damage can recall remote memories better than recent ones, a phenomeon called temporally-graded retrograde amnesia \cite{WINOCUR1990145, doi:10.1126/science.2218534, Rempel-Clower5233}. Thus, contrarily to the CLS theory, the SCT supports the hypothesis that the neocortex stores episodic memory transferred from the hippocampus. That would entail that the neocortex stores both semantic and episodic memories. 

However, the theory has been challenged after it has been found that the recall of remote memories is impaired following hippocampal lesions when the memories are truly episodic \cite{Viskontas2002-bh, Rosenbaum2000-vn, CIPOLOTTI2001151}. These findings suggest that the hippocampus is needed for recalling detailed episodic memory vividly. Thus, it is likely that in temporally-graded retrograde amnesia, remote memories could be recalled semantically, but not episodically. The multiple trace theory was proposed to account for these findings as an alternative to the SCT \cite{https://doi.org/10.1002/1098-1063(2000)10:4<352::AID-HIPO2>3.0.CO;2-D}. This view aligns with the CLS theory according to which the neocortex stores only semantic memory, and therefore episodic memories are transformed into semantic ones in the neocortex.  

\subsection{Artificial intelligence}

In artificial intelligence, episodic and semantic memories are traditionally implemented respectively by models like Hopfield networks \cite{doi:10.1073/pnas.79.8.2554} and Boltzmann machines \cite{10.7551/mitpress/5236.003.0011}. On the one hand, Hopfield networks are recurrent autoassociative memory (AM) models which store binary data points as attactors, so that they can be recovered by energy minimization when presenting noisy or partial versions. A modern, continuous version, called Modern Hopfield Networks (MHN), extends their memory capacities \cite{Krotov2023}. On the other hand, Boltzmann machines are energy-based generative models, which are considered as the predecessors of deep learning models, and especially Variational Autoencoders (VAE). The fast learning of individual examples in Hopfield networks and their recurrent structure is reminescent of the hippocampus (especially the hippocampal subfield CA3), whereas the slow learning of statistical structure in Boltzmann machines are reminescent of the neocortex. 

However, it has been shown that overparameterized autoencoders (AE) are also good AM: an AE trained to generate a particular data point stores it as an attractor \cite{DBLP:journals/corr/abs-1909-12362}. In addition, Salvatori et al. show that predictive coding networks (PCN), which have been proposed as a biologically plausible model of the neocortex \cite{friston2003learning}, outperform AEs and MHN in AM tasks \cite{salvatori2021associativememoriespredictivecoding}. 

Therefore, from a neuroscience perspective, it seems that semantic learning systems also have the ability to store individual experiences like episodic memory systems, even though these experiences have dense, overlapped representations which are acquired slowly. This has led researchers to relate them to the hippocampus, despite the differences in representation and storage. Indeed, due to the similarity in architecture and function, autoencoders have been used to model the hippocampus \cite{benna2021place, santos2021entorhinal}. Following the observation that the hippocampus is at the apex of the neocortical hierarchy \cite{barron2020prediction}, Salvatori et al. proposed that the top layer of their PCN could be viewed as the hippocampus, while lower layers would correspond to the neocortex. Therefore, the need for two separate learning systems put forward by the CLS theory seems to be questionned. 

In this work, we investigate the supposed episodic nature of the neocortex using a PCN, and asked how it can memorize individual examples through statistical learning. First, we recall the theory of predictive coding. Then, we derive the inference and learning updates for a 3-layer PCN. We also explain how this model can be mapped onto the brain and how replay and recall can be performed. In our experiments, we train the model on a subset of the MNIST dataset until all prediction errors are minimized and examine the learned model. We show that a PCN which is trained on a mini-batch of images is overfitted and can successfully recall them. When the same model is trained on the full dataset, recall performance deteriorates even though the model can successfully reconstruct the images, and recall seems mostly semantic. This study suggests that a limited number of episodes can be stored in a semantic system, but as this number grows, information about individual episodes is lost to the benefit of statistical knowlege. 

\section{Predictive coding}

Predictive coding is a model and an algorithm for inference and learning which has been mapped onto the cortical microcircuitry. It has its roots in studies of the visual cortex. Olshausen and Field show that learning a sparse code for natural images leads to the emergence of receptive fields which are similar to those found for simple cells in the mammalian primary visual cortex \cite{olshausen1996emergence}. They assume that an image with pixel intensity $I(x, y) $ at coordinates $x, y$ can be represented by a linear combination of basis functions, $\phi_i(x, y)$
\begin{equation}\label{eq:code}
I(x, y) = \sum_i a_i \phi_i(x, y).
\end{equation}
where the coefficients $a_i$ change from one image to the next. The goal of efficient coding is to find a set of basis functions $\phi_i(x, y)$ that forms a complete code (i.e. spans the entire image space) and results in the coefficient values $a_i$ being as statistically independant as possible over an ensemble of natural images. Sparse coding proposes to reduce statistical dependancies by ensuring that the representations are sparse, i.e. distributed among few coefficients. 

Considering coefficients $a_i$ as neuron activities and basis functions $\phi_i(x, y)$ as weights, sparse coding could be implemented as a two-layer predictive coding neural network with the image $I(x, y)$ in the input layer. Rao and Ballard make this model hierarchical by adding a third layer predicting the second, similarly to how the second layer predicts the first \cite{rao1999predictive}. Their three-level predictive coding network showed extra-classical receptive field effects, in addition to simple cell receptive fields, suggesting that these effects might emerge in the visual cortex from a hierachical and predictive strategy for encoding natural images. 

Friston goes further by proposing predictive coding as a general theory of cortical computation, given the similarity in architecture throughout the whole neocortex, with its hierarchical structure and reciprocal connections \cite{friston2003learning}. He adapts Rao and Ballard's model to a variational inference framework, extends it to $n$ layers and shows how it can be implemented in a biologically plausible way in the neocortex. He considers a conditionally independent hierarchical model (also called parametric empirical Bayes) with $n$ layers
\begin{equation*}\label{eq:empirical-bayes}
\forall i \in 1...n, p_{\theta_i}(\nu_i \mid \nu_{i+1}) = \mathcal{N}(\nu_i: G_i(\nu_{i+1}, \theta_i), \Sigma_i(\lambda_i))
\end{equation*}
where $\nu_1$ is the input variable, equal to the input $u$, and $\nu_2, ..., \nu_n$ are the latent variables. This model is a generative model, where each variable $\nu_i$ is caused by the variable in the level above $\nu_{i+1}$,  potentially reflecting the hierarchical causal structure of the world. $\theta_i$ and $\lambda_i$ parameterize the means and covariances of the likelihood at each level $i$, through functions $G_i$ and $\Sigma_i$. 
Recognition is assumed to be deterministic, such that
\begin{equation*} 
q_\phi(\nu_2, ..., \nu_n \mid u) = \prod_{i=2}^n \delta(\nu_i  - \phi_i)
\end{equation*}
where $\phi = (\phi_2, ..., \phi_n)$ is an estimate of the causes of the input $u$.

In a variational inference framework, inference of the latent variables and learning of the parameters correspond to the minimization of a lower bound to the negative likelihood, called variational free energy. The variational free energy for input $u$ is
\begin{align*}
\ell(u) &= - <\log p(u, \nu_2, ..., \nu_n)>_q\\
&= -\log p(u, \phi_2, ..., \phi_n)\\
&= -\log p(u \mid \phi_2) + \log p(\phi_2 \mid \phi_3) + ... + \\ 
&\qquad \log p(\phi_{n-1} \mid \phi_n) + \log p(\phi_n)\\
&= \frac{1}{2} \sum_{i=1}^{n-1} \left[ \xi_i^T \xi_i + \log |\Sigma_i| \right] + \log p(\phi_n) + \text{constant}
\end{align*}
where $\xi_i = \Sigma_i^{-\frac{1}{2}}(\phi_i - G_i(\phi_{i+1}, \theta_i))$ and the total variational free energy is
\[
F = <\ell(u)>_u.
\]

Implemented in a hierarchical predictive coding network, each layer $i$ consists of two types of neurons: units $\phi_i$ and error units $\xi_i$. Neurons with activity $\phi_i$ are predicted based on neurons in the layer above $i+1$ through backward connections $\theta_i$. Error neurons $\xi_i$ receive this prediction $G_i(\phi_{i+1}, \theta_i)$ from units in the level above and connections from the units $\phi_i$ being predicted to compute the prediction error. Choosing functions $\Sigma_i(\lambda_i) = (1 + \lambda_i)^2$ for the covariances, the prediction error becomes
\begin{align*}
\xi_i &= (1 + \lambda_i)^{-1}(\phi_i - G_i(\phi_{i+1}, \theta_i))\\
&= \phi_i - G_i(\phi_{i+1}, \theta_i) - \lambda_i \xi_i.
\end{align*}
and it can be seen that $\lambda_i$ are the strenghts of lateral connections among error units, which could serve to decorrelate them. 

The activities of the hidden neurons $\phi$ and connection strengths $\theta$ and $\lambda$ are obtained by performing gradient descent on the variational free energy using a form of the EM algorithm. 
\begin{enumerate}
\item E-step:
\[
\dot{\phi}_{i+1} = -\frac{\delta l(u)}{\delta\phi_{i+1}} = -\frac{\delta\xi_i^T}{\delta\phi_{i+1}}\xi_i - \frac{\delta\xi_{i+1}^T}{\delta\phi_{i+1}}\xi_{i+1}
\]
\item M-step:
\begin{align}
& \dot{\theta}_i = -\frac{\delta F}{\delta \theta_i} = - < \frac{\delta \xi_i^T}{\delta \theta_i} \xi_i >_u\\
& \dot{\lambda}_i = - \frac{\delta F}{\delta \lambda_i} = -<\frac{\delta\xi_i^T}{\delta\lambda_i}\xi_i>_u - (1 + \lambda_i)^{-1}
\end{align}
\end{enumerate}
We refer to the E-step as inference and the M-step as learning. In the E-step, the estimators $\phi_{i+1}$ are updated based on prediction errors $\xi_i$ carried forward fom the layer below, along with lateral interactions with the error units $\xi_{i+1}$. It can be seen that all calculations are local, and weight updates can be seen as Hebbian learning. More details of the proposed neuronal implementation can be found in Friston \cite{friston2003learning}. In later publications, Friston proposed that precision $\Sigma_i^{-1}$ is predicted by higher layers and mediates attention \cite{10.3389/fnhum.2010.00215}. Indeed, if precision $\Sigma_i^{-1}$ is low, prediction error $\xi_i = \Sigma_i^{-\frac{1}{2}}(\phi_i - G_i(\phi_{i+1}, \theta_i))$ is low, and will not influence the update of neuron activities and weights. 

\section{Methods}

We propose a predictive coding model to investigate how the learning of representations supports memory using the MNIST dataset. 

\subsection{Model}

Our model is a PCN with the 3-level hierarchical model
\begin{align*}
p_{\theta_1}(u \mid \nu_2) &= \mathcal{N}(u: \theta_1\nu_2, I) \\
p_{\theta_2}(\nu_2 \mid \nu_3) &= \mathcal{N}(\nu_2: f(\theta_2\nu_3), I) \\
p(\nu_3) &= \mathcal{N}(\nu_3: 0, I).
\end{align*}
where we use an identity covariance matrix for all layers and an identity activation function for the first layer following equation \ref{eq:code}. The variational free energy per input $u$ is 
\begin{equation*}
l(u) = \frac{1}{2}\xi_1^T\xi_1 +\frac{1}{2}\xi_2^T\xi_2 +\frac{1}{2}\xi_3^T\xi_3 + \text{constant}
\end{equation*}
where
\begin{align*}
\xi_1 &= u - \theta_1\phi_2 \\
\xi_2 &= \phi_2 - f(\theta_2\phi_3) \\
\xi_3 &= \phi_3.
\end{align*}
The inference and learning update rules can be derived by calculating the gradient of the variational free energy with respect to the estimators $\phi$ and parameters $\theta, \lambda$ respectively. With an identity function, the calculation is easy for the first layer since
\begin{align*}
\nabla_{\phi_2}l &= -\theta_1^T\xi_1 + \xi_2 \\
\nabla_{\theta_1}l &= -\xi_1 \phi_2^T.
\end{align*}
For the second layer, we define $l_2 = \frac{1}{2}\xi_2^T\xi_2$ and $l_3 = \frac{1}{2}\xi_3^T\xi_3$, so that we can express $\nabla_{\phi_3}l = \nabla_{\phi_3}l_2 + \nabla_{\phi_3}l_3$ where
\begin{align*}
\nabla_{\phi_3}l_3 &= \xi_3\\
\nabla_{\phi_3}l_2 &= \frac{\partial}{\partial \phi_3}\left[ \phi_2 - f(\theta_2\phi_3) \right] \xi_2 \\
&= -\frac{\partial}{\partial \phi_3} \left[ f(\theta_2\phi_3) \right] \xi_2 \\
&= -\theta_2^T \text{diag}\left[ f'(\theta_2\phi_3) \right] \xi_2
\end{align*}
resulting in
\[
\nabla_{\phi_3}l = -\theta_2^T \text{diag}\left[ f'(\theta_2\phi_3) \right] \xi_2 + \xi_3.
\]
The gradient with respect to $\theta_2$ can be obtained similarly to the backpropagation update rule using the chain rule, such that
\[
\nabla_{\theta_2}l = -\left[ \xi_2 \odot f'(\theta_2\phi_3) \right] \phi_3^T
\]
where $\odot$ is the element-wise product. 

\begin{figure}[tb]
         \centering
         \includegraphics[width=0.4\textwidth]{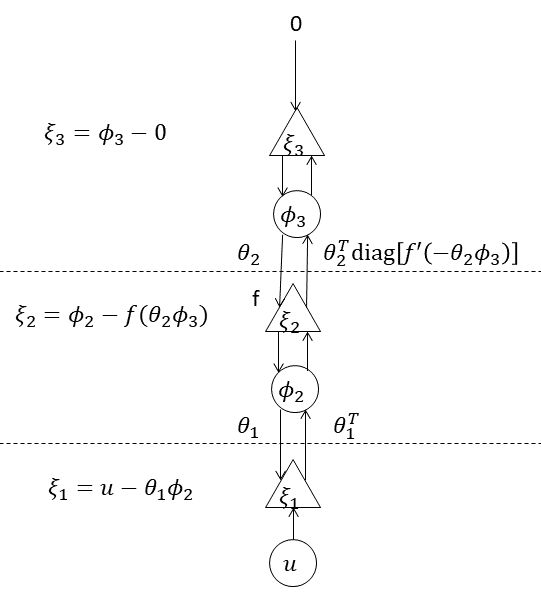}
         \caption{PCN with 3 layers. Each node corresponds to a layer corresponding to multiple neurons. Circle nodes correspond to units $\phi_i$ and triangle nodes correspond to error units $\xi_i$. Each connection between two layers corresponds to a fully connected network, with excitatory connections in the upward direction and inhibitory connections in the downward direction. We consider the upmost prediction to be 0, to have a standard normal distribution as prior. }
         \label{fig:model}
\end{figure}

Therefore, the inference and learning rules are
\begin{align}
\phi_2 &\leftarrow \phi_2 + \alpha (\theta_1^T\xi_1 - \xi_2) \label{eq:infer1} \\ 
\phi_3 &\leftarrow \phi_3 + \alpha (\theta_2^T \text{diag}\left[ f'(\theta_2\nu_3) \right] \xi_2 - \xi_3) \label{eq:infer2} \\ 
\theta_1 &\leftarrow \theta_1 + \beta <\xi_1 \phi_2^T>_u \label{eq:learn1}\\ 
\theta_2 &\leftarrow \theta_2 + \beta <\left[ \xi_2 \odot f'(\theta_2\phi_3) \right] \phi_3^T>_u \label{eq:learn2}
\end{align}
where $\alpha$ and $\beta$ are respectively the inference and learning rates. Figure \ref{fig:model} shows how the model can be seen as a neural network. 

\subsection{Replay and recall in PCN} \label{sec:replay}

Mapping our model to the visual pathway of the brain, the three layers would correspond respectively to the lateral geniculate nucleus (LGN), the visual cortex (VC) and the entorhinal cortex (EC). LGN is a structure in the thalamus which connects the retina to the visual cortex and EC is the main interface between the hippocampus and the neocortex. The input layer is mapped to the LGN and not the retina, because there is no backward connection from the LGN to the retina. When the LGN is presented with an image, the VC and EC converge to hierarchical representations of the image following the inference rules in equations \ref{eq:infer1} and \ref{eq:infer2}. It has been shown that low-level representations in the early visual cortex encode fine details of images, whereas high-level representations in associative areas, such as inferotemporal (IT) cortex, encode more global information about objects \cite{rao1999predictive}. In our model, representations $\nu_2$ in VC are predictive of the image $\nu_1$ in LGN, and the prediction $\theta_1 \nu_2$ corresponding to the image is called a reconstruction. 

Replay of episodic memories is triggered either by cues (during wakeful rest) or noise (during sleep) arriving at EC. Patterns in EC undergo pattern completion in the hippocampus and the recalled pattern in EC is then reinstantiated in the cortex, which is VC in our model. According to Barron \cite{barron2020prediction}, reinstated representations are 'protected' from ascending input or prediction errors from lower hierarchical levels. This could be done by setting the precision of the input layer $\Sigma_1^{-1}$ (or equivalently the prediction errors $\xi_1$) to 0. This will have the effect of cancelling the influence of the input on the above layer VC, which will only depend on EC. Therefore, replay of an input $u$ can be implemented with the following steps.
\begin{enumerate}
\item Infer $\nu_3$ using the inference rules from equations \ref{eq:infer1} and \ref{eq:infer2} until convergence.
\item Set $\nu_3$ to the value obtained in step 1 and $\xi_1$ to 0. 
\item Infer $\nu_2$ using equation \ref{eq:infer1} until convergence.
\item (optional) Update $\theta_2$ using equation \ref{eq:learn2}. 
\end{enumerate}
The prediction $\theta_1 \nu_2$ then corresponds to the replayed input. The last step is not necessary here since we do not consider the consolidation of episodic memories. 

Recall of episodic memories consists in the pattern completion of a corrupted (noisy or partial)  version. We will consider the more challenging recall (or AM) task when presenting a partial version. Following Salvatori et al. \cite{salvatori2021associativememoriespredictivecoding}, we initialize the input layer $\nu_1$ to the corrupted input and let the corrupted part converge with the update rule
\[
\phi_1 \leftarrow \phi_1 - \alpha \xi_1
\]
obtained using $\nabla_{\phi_1}l = \xi_1$ where $\phi_1$ is the estimate of $\nu_1$.
The other two layers $\nu_2, \nu_3$ are similarly inferred following equations \ref{eq:infer1} and \ref{eq:infer2} until convergence. The converged value for $\phi_1$ would correspond to the recalled input. 

\subsection{Experiments}

We implement the above PC model using the inference and learning rules in equations \ref{eq:infer1} to \ref{eq:learn2}. We use Pytorch to train the model on mini-batches using GPU, but without using the library's automatic differentiation and optimizers. In the inference phase, hidden neurons are initialized randomly and then updated for $T$ iterations according to the above inference rules. The final values are used to perform a single update on the weights $\theta$. We also implement Incremental PC (iPC), a more stable modification of PC where the weights are updated alongside the latent variables at every time step \cite{salvatori2024stablefastfullyautomatic}. 

Two experiments are carried out on a subset of the MNIST dataset with only digits 4 and 7. The dataset is split in a training set of size 10097 and validation and test sets, both of size 2010. Training and validation sets are used during the training, whereas the test set is used after training. In experiment 1, the PC model is trained on a single mini-batch of size 64 from the initial training set. In experiment 2, the iPC model is trained on all images from the training set. In both experiments, the model is trained for multiple epochs, until convergence of all prediction errors. Since convergence on the whole dataset in experiment 2 requires many weight updates, iPC is required for stability. Hyperparameters are chosen based on empirical trials informed by the literature \cite{Millidge2019, frieder2024bad, pinchetti2024benchmarkingpredictivecodingnetworks} and summarized in Table \ref{tab:hyperparam}. 

\begin{table}
\caption{Hyperparameter values for the two experiments}
\label{tab:hyperparam}
\centering
\begin{tabular}{lc*{8}{@{\hspace{-4pt}}c}}
\toprule
Model &	 PC && iPC \\
\midrule
Dimensions && [784, 35, 2] \\
Activation function $f$ && $\tanh$ \\
Batch size && 64 \\
Number of iterations $T$ && 50 \\
Inference rate $\alpha$ && 0.01 \\
Inference optimizer && SGD \\
Learning rate $\beta$ & $10^{-4}$ && $10^{-5}$ \\
Learning optimizer && Adam \\
\bottomrule
\end{tabular}
\end{table}

Except for the content of the training dataset, the two experiments are performed in the same way. During training, we keep track of the average sum in prediction errors in each layer for the training and validation sets. After training, we examine the learned models by looking at the weights $\theta_1$ and the reconstructions for a mini-batch from the training and test sets.  The same training mini-batch is replayed and the replayed inputs are visualized. Then, we plot the representations for the test set in each layer in 2 dimensions using t-SNE. Lastly, we perform an AM task by considering 10 images from the training set as episodic memories and recalling the bottom half of each image when presenting only the top half. Reconstruction, replay and recall are implemented following the methods in the previous subsection. 

\section{Results}

\subsection{Experiment 1: Training on a single minibatch}

\begin{figure}[tb]
     \centering
     \subfloat[Training set \label{fig:exp1_train_reconstructions}]{
         \includegraphics[width=0.48\textwidth]{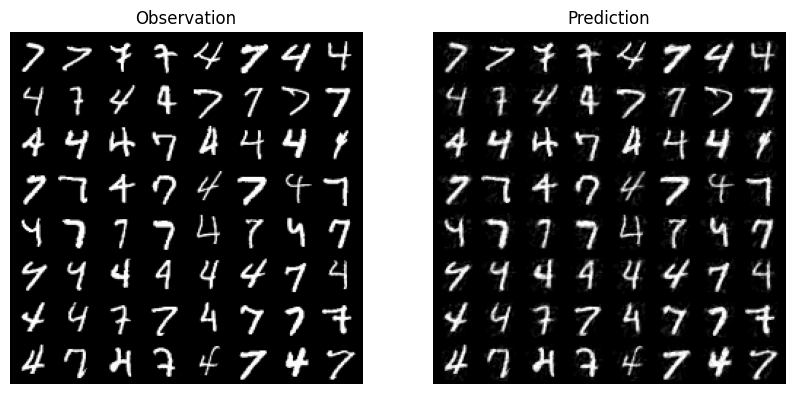}}
     \hfill
     \subfloat[Test set \label{fig:exp1_test_reconstructions}]{
         \includegraphics[width=0.48\textwidth]{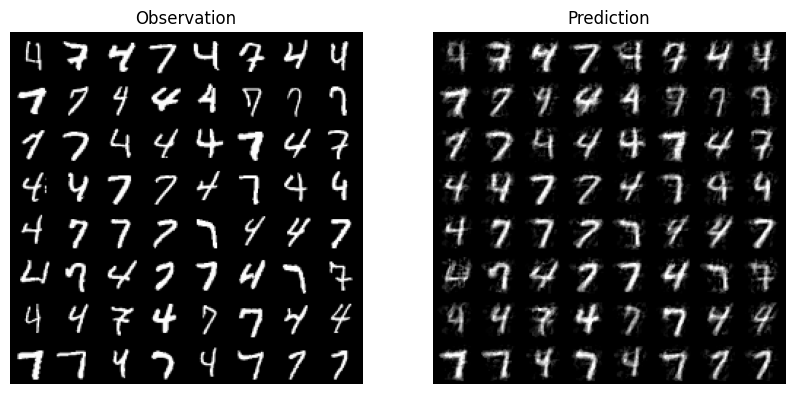}}
	\caption{Experiment 1: Reconstructions for a mini-batch of images from the training and test sets.}
	\label{fig:exp1_reconstructions}
\end{figure}

\begin{figure}[tb]
         \centering
         \includegraphics[width=0.20\textwidth]{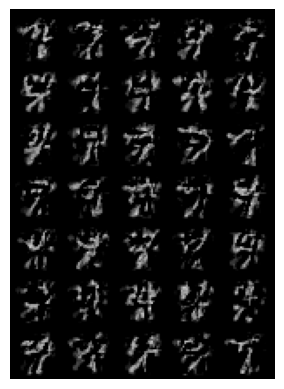}
         \caption{Experiment 1: Visualization of weights $\theta_1$ as 35 images of size $28 \times 28$.}
         \label{fig:exp1_weights}
\end{figure}

The prediction errors in the input layer converge to higher values for the validation set compared to the training set (not shown), suggesting that the model is overfitted to the minibatch. Indeed, as shown in Figure \ref{fig:exp1_reconstructions}, images from the test set are not reconstructed as well by the model compared to images from the training set. This is reflected in the weights $\theta_1$ of the model, which are shown in Figure \ref{fig:exp1_weights} as 35 images of size $28 \times 28$ that look like noisy combinations of examples of 4 and 7. 

\begin{figure}[tb]
         \centering
         \includegraphics[width=0.20\textwidth]{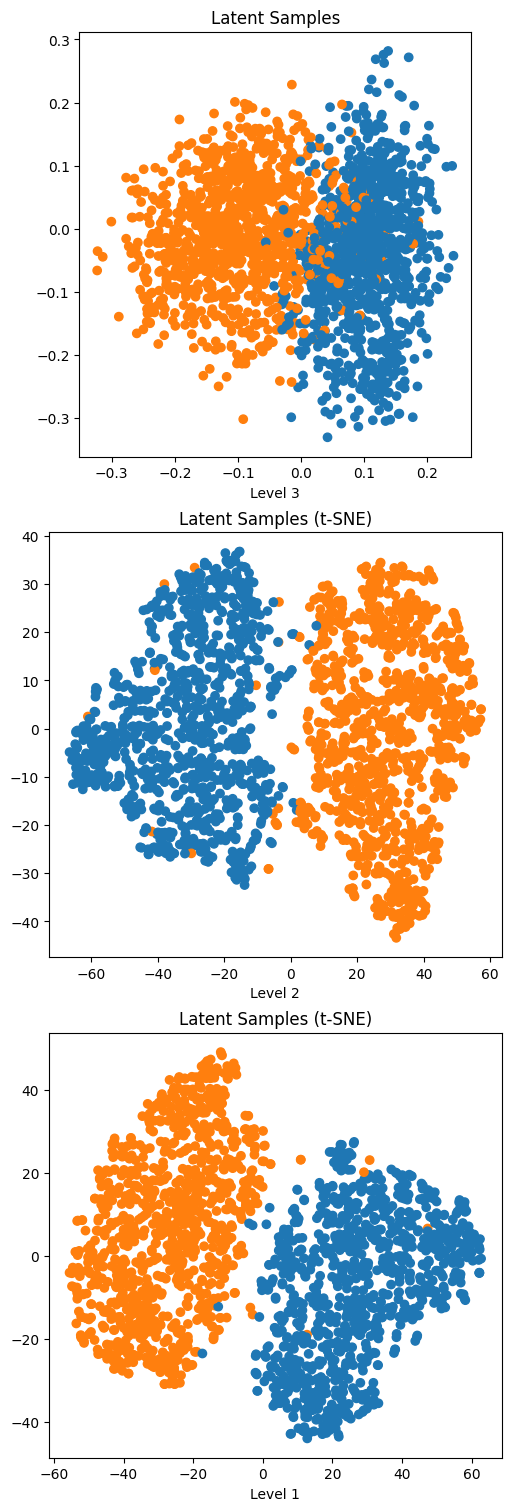}
         \caption{Experiment 1: Visualization of the hierarchical representations of the test data. Each image from the test set is represented as one data point in each of the three subplots, colored according to the class. Blue and orange represent digits 4 and 7 respectively. The first two levels (at the bottom of the figure), corresponding to $u$ and $\phi_2$, are visualized in two dimensions using t-SNE. }
         \label{fig:exp1_latents}
\end{figure}

\begin{figure}[tb]
         \centering
         \includegraphics[width=0.48\textwidth]{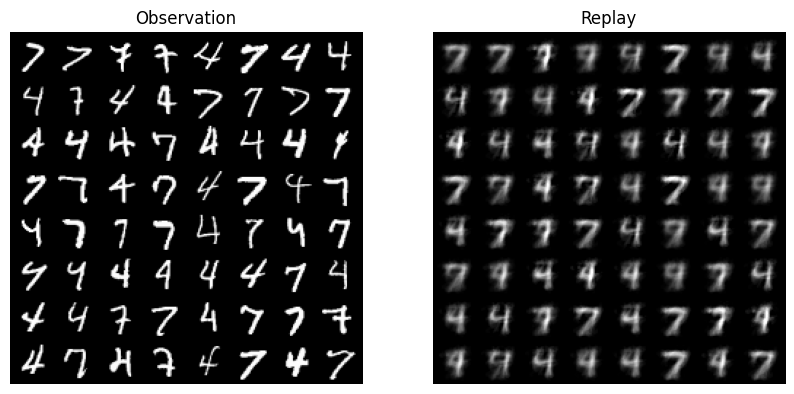}
         \caption{Experiment 1 : Replayed inputs (right) for a mini-batch of training images (left).}
         \label{fig:exp1_replay}
\end{figure}

The hierarchical representations of the test data are visualized in Figure \ref{fig:exp1_latents} using t-SNE to reduce the dimensionality of the first two levels to 2. The representations in the first two levels are well separated according to the class, suggesting that the images $u$ (bottom plot) and their causes $\nu_2$ lie in a 2-dimensional manifold corresponding to the two classes. Indeed, the estimates of the two-dimensional causes of $\nu_2$, $\nu_3$, are also disentangled according to the class. However, they overlap at the boundary between the two classes. This means that some digits might not be easily classified by our model. As a result, replayed inputs corresponding to these high-level representations might be ambiguous. This is what we see in Figure \ref{fig:exp1_replay}: some digits are misremembered when replayed, and often, the replayed inputs look like a combination of 4 and 7. The other images are replayed as the right class but the details of the original images are lost. This is expected because replayed images are obtained from high-level representations in EC, which are supposed to encode global information about the classes. 

\begin{figure}[tb]
     \centering
     \includegraphics[width=0.20\textwidth]{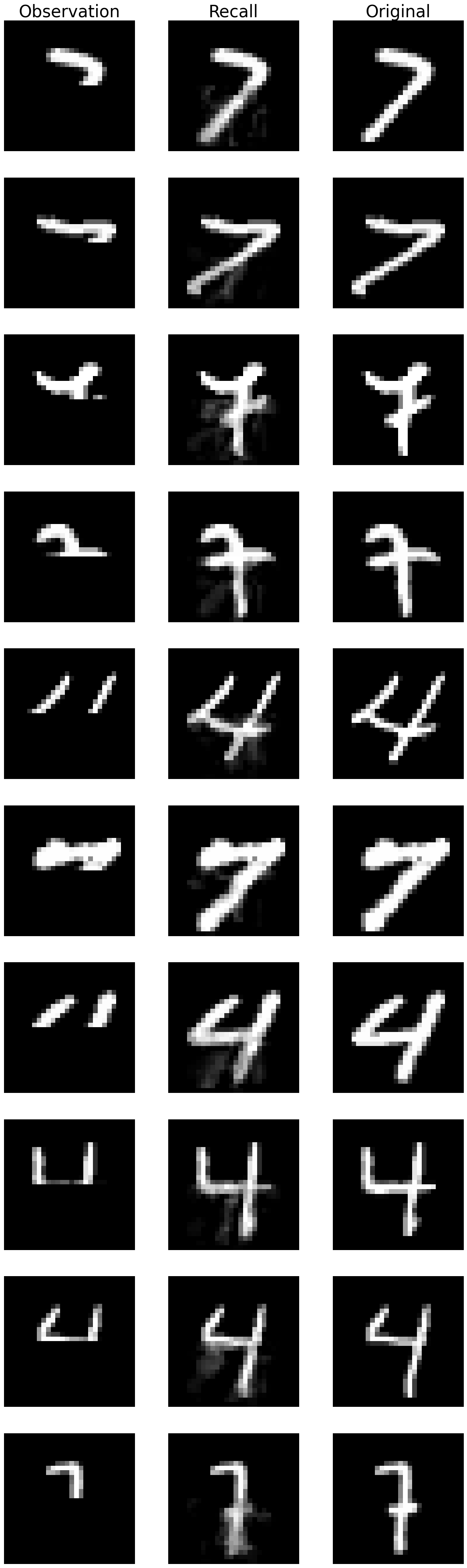}
	\caption{Experiment 1: AM task. Ten images from the training set (right column) are taken as episodic memories and the bottom half of the images is recalled (middle column) when presenting only the top half of the images (left column).}
	\label{fig:exp1_recall}
\end{figure}

Finally, the episodic memories from the training set were successfully recalled, as shown in Figure \ref{fig:exp1_recall}. 

\subsection{Experiment 2: Training on the full dataset}

The hierarchical representations of the test data and the replayed images are similar to those obtained in the previous experiment. However, there are major differences in the prediction errors and the recall. 
 
\begin{figure}[tb]
     \centering
     \subfloat[Train set \label{fig:exp2_train_reconstructions}]{
         \includegraphics[width=0.48\textwidth]{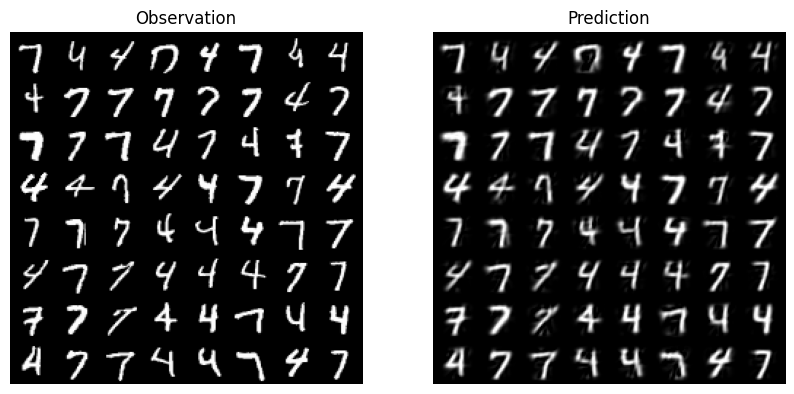}}
     \hfill
     \subfloat[Test set \label{fig:exp2_test_reconstructions}]{
         \includegraphics[width=0.48\textwidth]{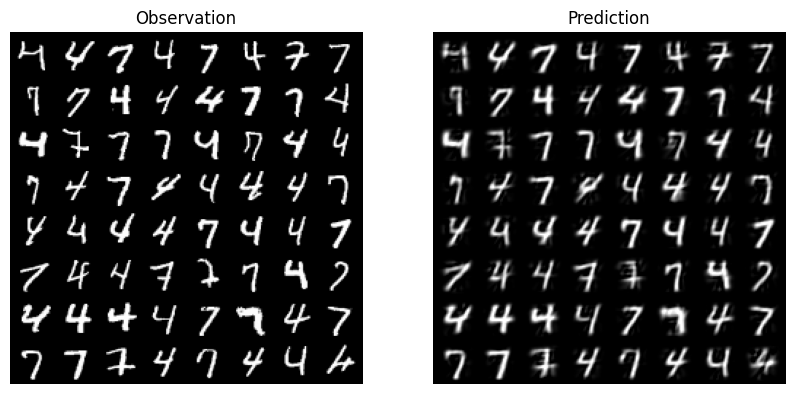}}
	\caption{Experiment 2 : Reconstructions for a mini-batch of images from the training and test sets.}
	\label{fig:exp2_reconstructions}
\end{figure}

\begin{figure}[tb]
         \centering
         \includegraphics[width=0.20\textwidth]{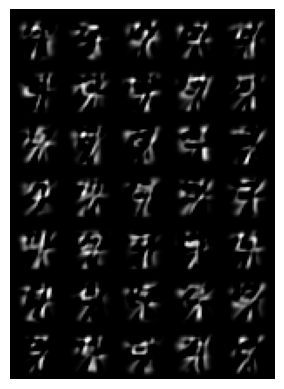}
         \caption{Experiment 2: Visualization of weights $\theta_1$ as 35 images of size $28 \times 28$.}
         \label{fig:exp2_weights}
\end{figure}

On the one hand, the training curves of the prediction errors for the training and validation sets overlap (not shown), suggesting a good fit. Low prediction errors in the input layer $\xi_1$ at the end of training for the training and validation sets suggest that the model is able to reconstruct well images of handwritten digits 4 and 7, whether they were seen during training or not. Indeed, we show in Figure \ref{fig:exp2_reconstructions} that images from the test set are reconstructed as well as those from the training set. This is due to weights $\theta_1$ being smoother than in the previous experiment, as shown in Figure \ref{fig:exp2_weights}, constituting a complete code for images of digits 4 and 7 from the MNIST dataset. 

\begin{figure}[tb]
     \centering
     \includegraphics[width=0.25\textwidth]{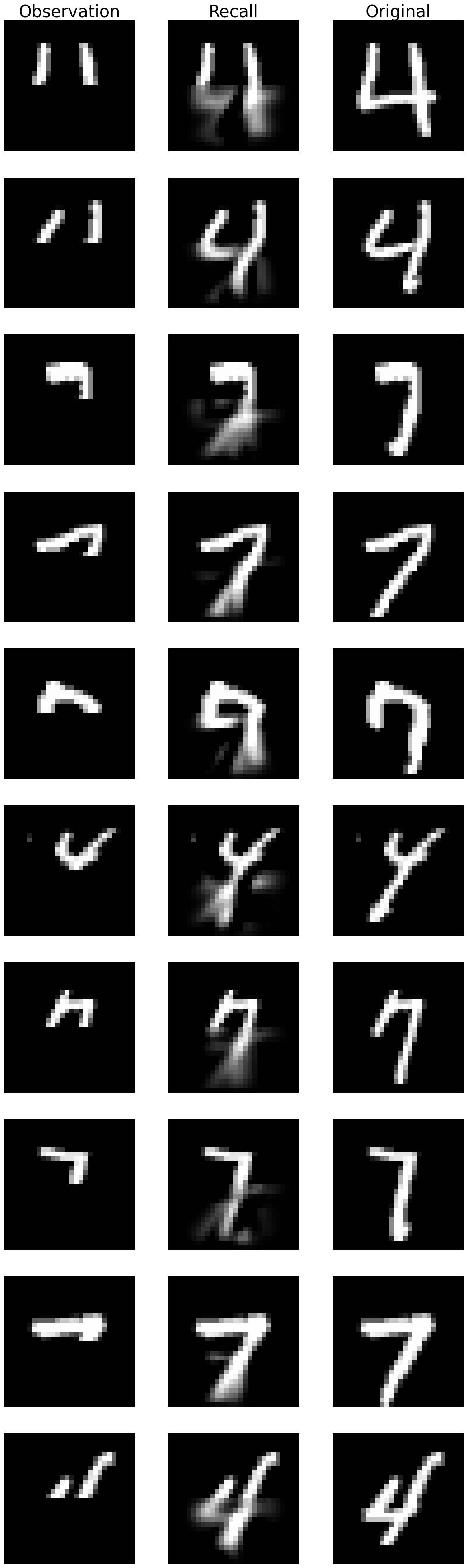}
	\caption{Experiment 2: AM task. Ten images from the training set (right column) are taken as episodic memories and the bottom half of the images is recalled (middle column) when presenting only the top half of the images (left column).}
	\label{fig:exp2_recall}
\end{figure}

On the other hand, Figure \ref{fig:exp2_recall} shows that the episodic memories are not recalled as well as in the previous experiment. Even though the images are recalled in a realistic way, the details of the original images seem to be lost and in some images, a bar appears in the middle of the digit, even though it is absent in the original images. This reflects semantic learning because many images of 4 and 7 in the training set have a bar in the middle. 

\section{Related Work}

Millidge implemented predictive coding in a similar way, using the same inference and learning rules, and examined the reconstructions and latent space of MNIST images \cite{Millidge2019}. However, these rules were not derived in the paper and the results were only shown for a two-layer PCN. The latest implementations used automatic differentiation with Pytorch to perform inference and learning \cite{salvatori2021associative, pinchetti2024benchmarkingpredictivecodingnetworks}. However, they are not trained until convergence of all prediction errors, but until performance on specific tasks has converged, because their goal is to apply predictive coding to AI tasks, and not to study the learning of representations in the neocortex. Salvatori et al. trained two two-layer PCNs with 1024 and 2048 hidden neurons on 50 images from the Tiny ImageNet dataset and succesfully recalled all of them when presenting only 1/8 of the images, similarly to our first experiment \cite{salvatori2021associative}. Pinchetti et al. studied the memory capacity of four-layer PCNs of increasing widths on a subset of the Tiny ImageNet dataset. They reported the MSE between the recalled and original images with model sizes $[12288, d, d, 512]$ where $d \in \{512, 1024, 2048\}$ and training datasets of sizes 50, 100 and 250. They also examine the reconstructions obtained by such a model when trained on an entire dataset of images, like in our second experiment. They compare the reconstruction errors, i.e. the MSE between the reconstructed and original images, of PC, iPC and VAE on different image datasets: MNIST, FashionMNIST, CIFAR-10 and CELEB-A. 

\section{Discussion}
In this paper, we proposed a predictive coding model of the neocortex to study the learning of representations and the formation of  memory. We showed that this model can be implemented as a neural network with local learning rules, and that it can be mapped onto the neocortical hierarchy. Using a subset of the MNIST dataset with only digits 4 and 7, we showed that the model can infer the two-dimensional structure of the dataset thanks to its hierarchical structure, as well as replay images considered as episodic memories. More importantly, we showed that our model can memorize a small training set by overfitting to it, but as the training set becomes larger, this ability is lost to the benefit of more general knowledge. As the neocortex is exposed to a vast quantity of sensory inputs, it likely cannot memorize individual episodes. Therefore, this work supports the CLS theory according to which the neocortex only supports semantic memory and episodic memory is supported by the hippocampus. This is likely due to the nature of representations learned by these two systems. On the one hand, dense, overlapping representations are learned slowly by the neocortex, and do not allow the recall of the specifics of individual examples. On the other hand, sparse, pattern-separated representions, which are learned rapidly by the hippocampus, support recall, but do not allow the extraction of statistical structure. Overall, this work contributes to a better understanding of learning in the hippocampus and the neocortex as well as their interactions from a neuroscience perspective, and also sheds light on the nature of generative models in AI. Future work could further examine the relationship between training set size and recall performance, with increasing model width and dataset complexity. 

\bibliographystyle{IEEEtran}
\bibliography{main}

\end{document}